# Development and Validation of SXI++ LNM Algorithm for Sepsis Prediction

*Dharambir Mahto[1], Prashant Yadav[1], Mahesh Banavar, P.hD[1,2], Jim Keany, MD[1,3], Alan T Joseph[1], Srinivas Kilambi, Ph.D., C.F.A.[1]*

[1.] *Sriya.AI, Atlanta, GA*
[2.] *Associate Professor, Department of ECE, Clarkson University, Potsdam, NY, USA*
[3.] *Office of the Chief Medical Officer, CMO CommonSpirit St Mary Medical Center, Long Beach, CA*

Author Contribution
(I) Conception and design: Srinivas Kilambi
(II) Administrative support: Prashant Yadav
(III) Provision of study materials or patients: Srinivas Kilambi, Mahesh Banavar & Jim Keany
(IV) Collection and assembly of data: Alan T Joseph
(V) Data analysis and interpretation: Dharambir Mahto, Alan T Joseph, Mahesh Banavar & Jim Keany
(VI) Manuscript writing: All authors
(VII) Final approval of manuscript: All authors

Correspondence to:
Srinivas Kilambi, P.hD, Chief Executive Officer, Sriya.AI, St. Marlo Atlanta, Georgia 30097, United States, sk@sriyaai.com

**Abbreviation list**

- **AUC** – Area Under the Curve
- **CI** – Confidence Interval
- **COMPOSER** – Conformal Multidimensional Prediction of Sepsis Risk
- **ML** – Machine Learning
- **NPV** – Negative Predictive Value
- **PPV** – Positive Predictive Value
- **SOFA** – Sepsis-related Organ Failure Assessment
- **SXI++** – Sriya Expert Index Plus

## Abstract:

**Background**
Sepsis is a life-threatening condition affecting over 48.9 million people globally and causing 11 million deaths annually. Despite medical advancements, predicting sepsis remains a challenge due to non-specific symptoms and complex pathophysiology. The SXI++ LNM model is a machine learning-based scoring system that refines sepsis prediction by leveraging multiple algorithms and deep neural networks. The COMPOSER model, a deep learning framework utilizing conformal prediction, aims to improve robustness in clinical applications. This study compares the predictive performance of SXI++ LNM and COMPOSER for sepsis prediction.

**Methods**
A dataset containing 1,552,210 rows with 43 columns was cleaned and refined to 964,355 rows and 14 key features for sepsis prediction. Data were sourced from ICU patients across three separate hospital systems, including two publicly available datasets from Kaggle and the Early Prediction of Sepsis from Clinical Data: The PhysioNet/Computing in Cardiology Challenge 2019. Eligibility criteria included patients with comprehensive vital signs and lab data. The study setting covered diverse clinical environments to ensure generalizability.

The SXI++ LNM model, utilizing a deep neural network, was trained and tested using multiple scenarios with different dataset distributions. The model's performance was assessed against unseen test data, and accuracy, precision, and the Area Under the Curve (AUC) were calculated. The SXI++ Large Numerical Model (LNM) was also compared with the COMPOSER model, evaluating their ability to predict early sepsis.

**Results**
The SXI++ LNM model outperformed COMPOSER in three use cases, achieving an AUC of 0.99 (95% CI: 0.98–1.00) compared to COMPOSER's AUC range of 0.91–0.95 (95% CI: 0.89–0.96). The model demonstrated a precision of 99.9% (95% CI: 99.8–100.0) and an accuracy of 99.99% (95% CI: 99.98–100.0), maintaining high reliability in real-world clinical settings.

**Conclusions**
The SXI++ LNM model offers a highly accurate and reliable approach for predicting sepsis within six hours before clinical diagnosis, outperforming the COMPOSER model. Its adaptability across diverse patient populations highlights its potential for enhancing early intervention and improving critical care outcomes.

**Highlight Box**

**1. Key Findings on Early Sepsis Prediction Models:**

- **SXI++ LNM:** Achieves near-perfect accuracy (~99%) with high precision (AUC 0.99), can potentially excel in early sepsis detection and improving patient outcomes.

**2. What is known and what is New:**

- **Known:** Early sepsis detection is critical, and advanced machine-learning models have been developed to address this challenge.
- **New:** This manuscript highlights SXI++ LNM's dynamic deep learning framework and its exceptional performance across various clinical datasets, can significantly enhance early sepsis prediction.

**3. Implications and Actions Needed:**

- **Improved Patient Care:** SXI++ LNM's high accuracy supports timely interventions, reducing sepsis mortality and improving adherence to sepsis management protocols.
- **Real-World Utility:** Consistent performance across diverse datasets positions SXI++ LNM as a reliable tool for healthcare providers.

- **Adoption of Predictive Technologies:** Encouraging healthcare systems to integrate advanced models like SXI++ LNM for better patient outcomes and reduced sepsis rates.

1. **Introduction**

    **Background**

1. Sepsis, a critical condition resulting from a severe immune response to infection, is among the leading global causes of death (1). Worldwide, around 48.9 million people are affected by sepsis annually, resulting in 11 million deaths (2, 3). It is also one of the costliest medical conditions to manage. Prior to the COVID-19 pandemic, sepsis treatment costs were estimated at $1.3 billion per year in Ontario, Canada, and $27 billion in the United States. Sepsis patients typically face hospital stays twice as long as those with other life-threatening conditions, and in-hospital mortality rates remain high at 20% (4, 5). Even after hospital discharge, survivors are at increased risk of death or suffer from a reduced quality of life (6, 7, 8 ) . Sepsis poses a significant challenge for global healthcare systems, with around 80% of cases present at the time of admission from the emergency department and the remainder occurring within hospital units (9). The primary issues in sepsis diagnosis include non-specific clinical symptoms, the lack of a biomarker with adequate sensitivity and specificity due to the condition's complex pathophysiology, and the fact that sepsis is a heterogeneous syndrome without a single identifiable cause, phenotype, or biological marker (9, 10). The latter group poses a special risk for delayed recognition. Hospital- onset sepsis patients are twice as likely to die, twice as likely to require mechanical ventilation, and have twice the number of ICU days. Given these challenges, early diagnosis and intervention—ideally within three hours as recommended by best practice guidelines—are crucial for improving patient outcomes (11, 12). Sepsis-associated respiratory failure and renal failure are associated with higher mortality, increased cost and increased requirements for intensive care (13). Readmission rates for sepsis are 1.52 times higher than other hospital discharge diagnosis (14). The cost is 5 times that of sepsis present on admission. This makes ongoing, passive surveillance of patients with early warning of sepsis onset critically importance (15).

2. The COMPOSER algorithm is a deep learning model specifically designed for early sepsis prediction, with a focus on reducing false alarms and ensuring reliable predictions. COMPOSER incorporates three key modules: (1) a weighted input layer that prioritizes recent clinical measurements, mimicking a clinician's approach, (2) a conformal prediction network to assess data reliability and identify cases outside the model's conditions for use, and (3) a sepsis predictor that generates a probability score for sepsis onset. The weighted input layer scales clinical variables based on the time since they were last measured, ensuring that more recent data carries greater weight. This design prevents the model from exploiting institutional biases, such as the frequency of measurements correlating with disease severity, thus enhancing generalizability. The conformal prediction module ensures predictions are only made when new data is statistically similar to the training data, using trust sets and hypothesis testing to determine conformity. These innovative modules work together to provide accurate predictions while maintaining reliability, especially in cases of data distribution shifts. By integrating these components, COMPOSER ensures that predictions are made only under reliable conditions, addressing a key challenge in dynamic and heterogeneous clinical environments. This study compares the COMPOSER algorithm with the SXI++ LNM to evaluate their respective strengths and limitations in early sepsis detection.

3. A recent study introduced COMPOSER (Conformal Multidimensional Prediction of Sepsis Risk), a deep learning model designed to enhance early sepsis prediction. COMPOSER is unique in its ability to flag unfamiliar cases—resulting from erroneous data, missing values, or distributional shifts—as indeterminate, thus avoiding unreliable predictions. Trained and validated using six patient cohorts, including 515,720 patients from ICUs and EDs across two U.S. healthcare systems, it achieved high AUC scores (ICU: 0.925–0.953; ED: 0.938–0.945. The model demonstrated clinical utility by providing early warnings— flagging patients approximately 12.2 hours before antibiotic administration in ICUs and 2.1 hours in EDs— allowing timely interventions and prioritization of high-risk patients (16). Additionally, a quasi-experimental study at UC San Diego Health System assessed COMPOSER's impact, revealing a 1.9% absolute reduction in in-hospital sepsis mortality (17% relative decrease), a 5.0% absolute increase in sepsis bundle compliance (10% relative increase), and a 4% reduction in SOFA (Sequential Organ Failure Assessment) scores 72 hours post-sepsis onset underscore COMPOSER's potential to improve patient outcomes while integrating seamlessly into clinical workflows.

SOFA: It is a scoring system used in medical settings to assess and track the severity of organ dysfunction in critically ill patients, particularly in intensive care units (ICUs). It helps clinicians evaluate the extent of a patient's organ failure and predict outcomes, such as the likelihood of survival (17).

4. Earlier treatment of sepsis leads to decreased mortality. Epic is an electronic medical record providing a predictive alert system for sepsis, the Epic Sepsis Model (ESM) Inpatient Predictive Analytic Tool. External validation of this system is lacking. This study aims to evaluate the ESM as a sepsis screening tool and determine whether an association exists between ESM alert system implementation and subsequent sepsis-related mortality (18).

**Rationale and Knowledge Gap**

5. Sepsis remains a leading cause of mortality globally, with existing detection models such as SOFA and SIRS exhibiting limited sensitivity and specificity. Machine learning-based approaches like COMPOSER have demonstrated potential, yet challenges persist in adapting to complex, heterogeneous datasets and maintaining predictive accuracy in diverse clinical settings. Despite advancements, no model has successfully combined high adaptability with near-perfect accuracy, particularly in real-world scenarios involving imbalanced datasets and unseen patient populations. Moreover, existing models often fail to provide actionable insights for clinical decision-making, limiting their utility in critical care environments. To address these challenges, predictive models like the SXI++ Large Numerical Model (LNM) and the COMPOSER have been developed. The SXI++ model transforms complex, multi-dimensional problems into simpler, actionable solutions through a dynamic scoring system. By combining 5–10 machine learning algorithms, it derives a weighted composite score reflecting key features and their correlations with early sepsis. This scoring system is further optimized through a proprietary deep neural network, enabling precise and adaptable predictions for early sepsis detection.

6. The COMPOSER model, on the other hand, leverages a conformal prediction framework to reduce false alarms, ensuring reliable predictions even in the presence of data distribution shifts. While COMPOSER is widely used and available for real-world clinical applications, its limitations in achieving near-perfect accuracy and its focus on conformal predictions make it an ideal benchmark for evaluating the SXI++ model. The SXI++ model, with its superior accuracy and flexibility across diverse datasets, offers a promising alternative that can advance sepsis detection strategies and improve patient outcomes. This study aims to compare the performance of these two models, emphasizing SXI++'s advanced capabilities in dynamic scoring, deep learning optimization, and adaptability to diverse clinical environments.

**Objective**

7. Previous studies predominantly employed machine learning techniques to address the challenge of early sepsis prediction, particularly in identifying high-risk patients in clinical environments. However, these studies often struggled to interpret and explain the influence of various clinical and physiological factors on sepsis predictions, to assess improvements in early sepsis detection over short-term, mid-term, and long-term periods, or to implement a comprehensive model capable of predicting sepsis without requiring extensive prior training. Additionally, there was a lack of integration of Large Numerical Models (LNMs), which could utilize extensive patient data from different clinical settings to predict sepsis onset more accurately in specific populations. By incorporating such LNMs, which aggregate data on vitals, lab results, medical history, and treatments, the potential for more precise early sepsis prediction across diverse patient groups is greatly enhanced, providing a more holistic understanding of sepsis vulnerability. To address these gaps, this study focused on: (i) evaluating the performance of the SXI++ model as a multivariate scoring system to predict early sepsis as a binary classification problem, (ii) enhancing the SXI++ scoring methodology using the Proprietary Deep Neural Network algorithm and correlating it with early sepsis prediction rates over time, and (iii) employing a targeted decision tree framework to interpret the most effective pathways leading to both early and delayed sepsis predictions, thereby offering evidence-based recommendations for improving early sepsis detection strategies and reducing overall sepsis incidence in vulnerable populations (iv) comparing the results from COMPOSER algorithm and its prediction ability and its inability to provide scientific suggestions for Sepsis control. We present this article in accordance with the TRIPOD+AI reporting checklist.

This study aims to develop and validate the SXI++ LNM model for early sepsis prediction, specifically designed to assist emergency and critical care doctors in making timely and accurate clinical decisions to improve patient outcomes. We present this article/case in accordance with the TRIPOD reporting checklist (available at https://jmai.amegroups.com/article/view/10.21037/jmai-24-393/rc).

## 2. Methods:

### 2.1 Methodology Overview

Since the research involved secondary analysis of anonymized data and did not include direct interaction with or intervention in human subjects, an ethics board review was not required. The study did not involve sensitive or personal data that would necessitate informed consent. The data used in this study were obtained from three geographically distinct U.S. hospital systems, each utilizing different electronic medical record systems: Beth Israel Deaconess Medical Center (hospital system A), Emory University Hospital (hospital system B), and a third, unidentified hospital system (hospital system C). The datasets were collected over the past decade with Institutional Review Board (IRB) approvals from the respective institutions. The data sources included routine care records, covering a comprehensive range of patient demographics, vital signs, and laboratory values. This ensured the inclusion of representative patient populations across diverse clinical settings. The study period encompassed the collection of records over the past decade, enabling a robust analysis of sepsis onset and prediction.

The raw dataset utilized in this study initially contained 1,552,210 rows and 43 columns. To improve data quality, we performed a cleaning process where we removed columns with more than 40% missing values. This resulted in a refined dataset comprising 964,355 rows and 14 columns. The cleaned dataset was subsequently divided into various subsets to evaluate the model's performance under different scenarios. In each scenario, two sets of data were designated as "unseen data" to validate the algorithm's effectiveness. Each of the unseen test datasets contained 50,000 rows, with 1,000 positive sepsis cases and 49,000 negative sepsis cases, allowing us to explore different distributions of the target variable.

In the context of model training, for the first use case, the training dataset comprised 50,404 rows, which included 15,404 records indicating positive sepsis cases and 35,000 records indicating no sepsis. The unseen test datasets, each containing 50,000 rows with 1,000 sepsis cases and 49,000 non-sepsis cases, were used to assess the model's performance. The second use case had a training dataset of 50,000 rows, containing 1,000 positive sepsis cases and 49,000 negative cases, with both unseen test datasets similarly structured. In the third use case, the training dataset included 10,000 rows, consisting of 3,000 positive sepsis cases and 7,000 negative cases, while the unseen test datasets had 50,000 rows, with 900 positive cases and 49,100 negative cases.

During the training phase, the Large Numerical Model (LNM) algorithm was used on the specified training datasets, ensuring a robust learning process. Additionally, we parallelly conducted AutoML prediction for each use case to compare the performance with SXI++ LNM and assess the ability of SXI++ LNM to perform across different distributions of sepsis detection.

During the evaluation phase, the trained model was validated using the two separate unseen datasets for each use case. To assess the model's generalization capabilities to new data, performance metrics such as accuracy, precision, and the area under the curve (AUC) were calculated. We also compared the SXI++ LNM with the COMPOSER algorithm based on model performance, providing insights into their respective strengths and weaknesses. The SXI++ LNMs are designed to deliver immediate, mid, and long-term early sepsis detection rates, along with actionable scientific recommendations that explain the factors influencing early sepsis detection. This comprehensive analysis underscores the model's ability to adapt and provide accurate predictions, ensuring its reliability in real-world scenarios and enhancing clinical decision-making.

## 2.2 Dataset Description

This study utilized a dataset comprising 1,552,210 rows and 43 columns to enhance early sepsis prediction. It includes vital signs such as heart rate, oxygen saturation, temperature, and blood pressure measurements, alongside key laboratory values like bicarbonate, lactate, and electrolyte levels. These data points are crucial for identifying deviations indicative of sepsis onset. In addition to vital signs and laboratory metrics, the dataset features demographic information, including age, gender, and ICU unit identifiers, enriching the analysis. The entire feature list is in (Table 1).

Table: 1 List of Initial features

| SL.NO | Features | SL.NO | Features |
|---|---|---|---|
| 1 | Hour | 23 | Glucose |
| 2 | HR | 24 | Lactate |
| 3 | O2Sat | 25 | Magnesium |
| 4 | Temp | 26 | Phosphate |
| 5 | SBP | 27 | Potassium |
| 6 | MAP | 28 | Bilirubin_total |
| 7 | DBP | 29 | TroponinI |
| 8 | Resp | 30 | Hct |
| 9 | EtCO2 | 31 | Hgb |
| 10 | BaseExcess | 32 | PTT |
| 11 | HCO3 | 33 | WBC |
| 12 | FiO2 | 34 | Fibrinogen |
| 13 | pH | 35 | Platelets |
| 14 | PaCO2 | 36 | Age |
| 15 | SaO2 | 37 | Gender |
| 16 | AST | 38 | Unit1 |
| 17 | BUN | 39 | Unit2 |
| 18 | Alkalinephos | 40 | HospAdmTime |
| 19 | Calcium | 41 | ICULOS |
| 20 | Chloride | 42 | SepsisLabel |
| 21 | Creatinine | 43 | Patient_ID |
| 22 | Bilirubin_direct | | |

Finally, the dataset includes a Sepsis Label as the target variable indicating whether patients were diagnosed with sepsis within a defined timeframe. Specifically, a value of 1 indicates that a patient met sepsis criteria at least 6 hours before the clinical prediction of sepsis, while a value of 0 signifies the absence of sepsis. This label is critical for evaluating the effectiveness of predictive models and understanding the timing of sepsis onset in relation to clinical interventions. Overall, the rich diversity of information in this dataset provides an invaluable resource for advancing the early detection of sepsis and improving patient outcomes in critical care settings.

The distribution of the target variable, "Sepsis Label," which classifies patients into two categories: "No Sepsis" (label 0) and "Sepsis" (label 1). Out of the total dataset, 946,951 cases (98.2%) were labeled as "No Sepsis," while 17,404 cases (1.8%) were identified as "Sepsis." This highlights a significant class imbalance, with a much larger proportion of non-sepsis cases compared to sepsis cases, which is a common challenge in medical predictive modeling.

## 2.3 Data Preprocessing

The dataset underwent a comprehensive analysis to handle missing values, leading to the removal of features with more than 40% null values. This meticulous preprocessing ensured that only the most relevant features were preserved for analysis. For the remaining features, missing values were imputed using appropriate statistical techniques—mean for continuous variables and mode for categorical variables. As a result of this data cleaning process, the final dataset was refined to 964,355 rows and 14 columns, each representing key features essential for predictive modeling in early sepsis detection. The 14 retained features include Hour, HR, O2Sat, SBP, MAP, DBP, Resp, Age, HospAdmTime, ICULOS, SepsisLabel, Patient_ID, Gender_0, and Gender_1.

## 2.4 SXI++ Large Numerical Model Framework

The SXI++ is a dynamic scoring system that transforms complex, multi-dimensional problems into a simpler, two-dimensional solution. It derives a weighted composite score from 5-10 machine learning algorithms, reflecting the most significant features and their impact on early sepsis. By iteratively adjusting these weights through a proprietary deep neural network, SXI++ enhances its correlation with early sepsis, improving prediction accuracy. This score is then used to optimize decision-making, such as reducing patient sepsis rates or improving overall healthcare system.

**Preprocessing and Normalization**

The SXI++ scoring mechanism begins with the preprocessing and normalization of the input dataset. This critical stage standardizes the data to ensure that all features are on a comparable scale, preventing any single feature from unduly influencing the final score. The normalization process takes into account the minimum and maximum values of each feature and adjusts them based on their correlation with the target variable. Features that are positively correlated are normalized by dividing the feature value by the maximum value, while negatively correlated features are normalized by subtracting the feature value from the maximum value and then dividing by that maximum value. This results in a normalized dataset that accurately reflects the relative importance and scale of each feature.

## 2.5 Statistical Analysis:

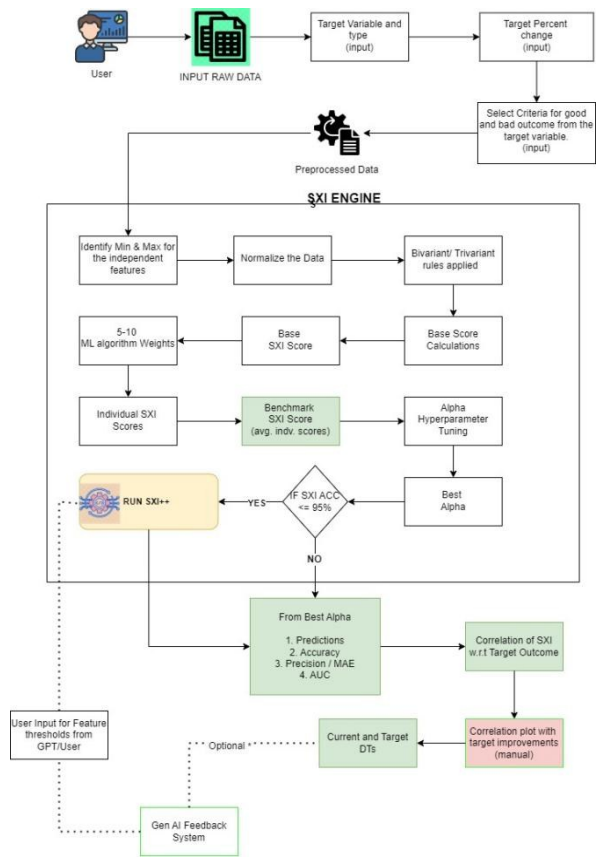

Fig: 1 SXI++ Framework, Score + Proprietary Deep Neural Network algorithm + Correlate = Improve. SXI

Bivariate Correlation Analysis

After normalization, the data undergoes a bivariate correlation analysis in (Fig 1), where pairwise correlations between all features are calculated. This step generates bi-variate correlation weights, which represent the average correlation of each feature with all others in the dataset. These weights are essential for understanding the interdependencies among features and their collective influence on the target variable. By using these correlation weights, the SXI++ mechanism identifies the most impactful features and adjusts their contributions to the final score accordingly.

Base SXI++ Score Calculation

Following the correlation analysis in (Fig 1), the SXI++ scoring system calculates the individual base SXI++ scores. This involves computing a weighted sum of the normalized feature values, where the weights are derived from the bi-variate correlation analysis. The individual scores are then averaged to obtain the base SXI++ score, which serves as a benchmark for further analysis. Additionally, binary labels, known as Base SXI++ Flags, are created by comparing each individual score with the base SXI++ score, categorizing data points into two groups based on their relative performance (19).

Lasso Regression Adjustment

The SXI++ mechanism in (Fig 1) then utilizes a Lasso regression model to update the min-max mapping of each feature. In this step, the model is fitted to the normalized data and Base SXI++ Flags, allowing for adjustments to feature importance based on the Lasso coefficients. Features with positive coefficients have their max values updated, while those with negative coefficients have their min values adjusted. This iterative process ensures that the feature mappings accurately represent their true influence on the target variable, leading to a more refined set of normalized data.

### Composite Weight Calculation and Final SXI++ Score

Finally, the SXI++ scoring in (Fig 1) mechanism integrates multiple machine learning algorithms to derive composite weights for the features. These algorithms include Complement Naïve Bayes, XGBoost, Mutual Information, Lasso Regression, and Principal Component Analysis (PCA). The weights generated by these algorithms are combined to compute the final SXI++ score, ensuring robustness and reliability. Features with non-zero weights from each algorithm are retained, and their contributions are adjusted based on their significance. The final composite weights are then used to calculate individual SXI++ scores, which are averaged to produce the final benchmark SXI++ score.

### Benchmark Analysis

Benchmark analysis of the baseline score involves extracting key metrics such as the Average SXI++ Score (Benchmark Score), the percentage of positive and negative outcomes, and the distribution of outcomes relative to the average SXI++ score (SXI++ Distribution w.r.t Outcomes). This analysis provides valuable insights into the performance of the SXI++ system in differentiating between favourable and unfavourable outcomes.

### Comparison with Delineation

To achieve maximum delineation, iterative improvements are made by comparing the initial SXI++ distribution with subsequent iterations. This process ensures that the delineation between different outcome groups is optimized over time.

### Working of Proprietary Deep Neural Network

The proprietary deep neural network architecture integrates independent features ($X_1, X_2, X_3, X_4...X_i$) and a target variable ($y$), along with updated base SXI++ scores and feature importance weights derived from common top features in the machine learning model. Consider a dataset with features labelled Feat1 to Feat15; extra weightage is given to the top 5 common features, with importance based on their frequency of occurrence (20, 21)

Initially, each feature is assigned a baseline weight of 1, and this weight is increased based on how frequently the feature appears in the top 5 common features. This approach ensures that more frequently occurring features are given higher importance, thereby enhancing their role in the analysis. The Custom Kernel Initializer function is utilized to modify the weight initialization process, giving extra weightage to the most important features based on their calculated importance.

The custom kernel initialization strategy involves calculating the effective input size using feature importance, followed by employing the Xavier/Glorot initialization method to generate random weights that account for custom feature weightage. The deep neural network's architecture is then configured with an 80/20 train-test split and fine-tuned using Bayesian optimization to determine the best hyperparameters, including neurons, activation functions, optimizers, learning rates, batch sizes, and epochs.

This optimization process is supported by stratified k-fold cross-validation to ensure the model's performance and generalization are robust, particularly when data is limited. The final model is compiled with a chosen optimizer and loss function, such as binary cross-entropy, and trained using the best hyperparameters identified through Bayesian optimization. The weights of the first five layers are then used to generate new SXI++ scores, ensuring that the model's predictions are based on the most refined feature importance.

### Iterative Weight Calibration System

The iterative weight calibration system aims to improve the SXI++ score and accuracy through a series of weight adjustments. The process begins with an initial assessment, where the current weights are used to calculate the SXI++ score and class delineation accuracy, serving as a benchmark for future comparisons. The system first evaluates whether any improvements can be made without adjusting the weights. If no improvement is found, it proceeds with positive weight adjustments, ranging from 0% to 100%, calculating the new SXI++ score and accuracy at each step. If positive adjustments show improvements, further adjustments are made beyond 100% until no additional gains are observed.

If positive adjustments do not yield improvements, the system then explores negative weight adjustments from 0% to -100%, following a similar process to identify the maximum delineation and accuracy gains within this range. Further negative adjustments are made if necessary, continuing beyond -100% until no further gains are detected. If neither positive nor negative adjustments improve the SXI++ score and accuracy, the scenario is considered non-improving. In this case, a new set of weights is generated, giving additional weightage to the top 5 most important features identified in the hidden layers of the neural network. These weights are then adopted as the new benchmark for future comparisons. This iterative process is repeated multiple times, refining the system by adjusting weights, applying penalties, and rewarding positive outcomes until the optimal delineator is achieved, ensuring that the SXI++ scores are continually improved.

**Model Training and Evaluation**

The model training process starts with hyperparameter tuning, specifically focusing on the alpha parameter. This involves adjusting the SXI++ score by varying the alpha value between 0.5 and 1.5 in increments of 0.1. During each iteration, the SXI++ score is recalculated by multiplying the current SXI++ score with the alpha value, enabling fine-tuning to optimize the model's performance.

$$SXI++_{new} = \propto * SXI++_{Current}$$

where:

- $\propto$ is the tuning parameter, ranging from 0.5 to 1.5 in increments of 0.1.
- $SXI++_{Current}$ is the Benchmark SXI++ score.
- $SXI++_{new}$ is the new SXI++ score after applying the alpha adjustment.

The dataset is divided into three parts: 70% for training, 20% for testing, and 10% for validation. This strategic split ensures the model is well-trained while also providing sufficient data for testing and validation, reducing the risk of overfitting and enabling an accurate assessment of model performance. During training, the optimized SXI++ scores are used as a "Super Feature," playing a crucial role in enhancing the model's predictive accuracy.

To thoroughly evaluate the model's performance, various metrics are employed. Accuracy measures the overall correctness of the model's predictions, while precision focuses on the proportion of true positive results among all positive predictions. The confusion matrix offers a detailed breakdown of true positive, true negative, false positive, and false negative outcomes, providing deeper insights into the model's classification abilities. Additionally, the Area Under the Curve (AUC) is utilized to assess the model's effectiveness in distinguishing between classes, serving as a robust indicator of its predictive power.

$$Accuracy = \frac{TP + TN}{TP + TN + FP + FN}$$

$$Precision = \frac{TP}{TP + FP}$$

Where:

- TP is the number of true positives.
- TN is the number of true negatives.
- FP is the number of false positives.
- FN is the number of false negatives.

The AUC for the Receiver Operating Characteristic (ROC) curve measures the classifier's ability to distinguish between positive and negative classes. It is calculated based on the True Positive Rate (TPR) and False Positive Rate (FPR), with the ROC curve plotting TPR against FPR at different threshold settings. TPR and FPR are defined as follows:

$$TPR = \frac{TP}{TP + FN}$$

$$FPR = \frac{FP}{FP + TN}$$

$$AUC \approx \sum_{i=1}^{n-1} \left(\frac{TPR_{i+1} + TPR_i}{2}\right)(FPR_{i+1} - FPR_i)$$

Where:

- $TPR_i$ and $TPR_{i+1}$ are the true positive rates at consecutive thresholds.
- $FPR_i$ and $FPR_{i+1}$ are the false positive rates at consecutive thresholds.

**Actionable Insights for Early Sepsis Detection**

In the decision tree model for early sepsis detection, clinical features that contribute positively to the likelihood of sepsis are assigned positive weights, while those that indicate a lower likelihood of sepsis are assigned negative weights. To implement strategies for reducing the risk of misclassification as sepsis, users specify the percentage increase for positive weighted features (which elevate the risk of sepsis) and the percentage decrease for negative weighted features (which help indicate the absence of sepsis). During the data transformation process, these adjustments are applied to each observation: positive weighted features are increased by the specified percentage, while negative weighted features are decreased, enhancing the dataset's overall accuracy for detecting sepsis (22).

If the early sepsis detection model shows a positive correlation between SXI++ scores and the likelihood of sepsis, the strategy for adjusting positive weighted features that indicate sepsis is:

$$Adjusted\ Feature\ Value = Orginal\ Feature\ Values * (1 + Percentage)$$

For features that suggest the absence of sepsis (negative weighted features), the strategy is:

$$Adjusted\ Feature\ Value = Orginal\ Feature\ Values * (1 - Percentage)$$

Conversely, if a negative correlation exists between SXI++ scores and sepsis (indicating that certain features may suggest the absence of sepsis), the weights are reversed. In this scenario, features that positively impact the indication of sepsis are assigned positive weights, while those that signify no sepsis receive negative weights. Therefore, for positive weighted features (which indicate sepsis when their values increase), the strategy involves a percentage decrease, while negative weighted features (which suggest no sepsis when their values increase) require a percentage increase.

For positive weighted features (which indicate sepsis when their values increase), the strategy is:

$$Adjusted\ Feature\ Value = Orginal\ Feature\ Values * (1 - Percentage)$$

For negative weighted features (which indicate no sepsis when their values increase), the strategy is:

$$Adjusted\ Feature\ Value = Orginal\ Feature\ Values * (1 + Percentage)$$

Once the dataset has been adjusted using the specified transformations, it is employed to train a Random Forest model with a maximum depth of 4. The Random Forest model is an ensemble learning algorithm based on decision trees, which constructs multiple decision trees during training and aggregates their outputs to improve classification accuracy while minimizing overfitting. In the context of a Random Forest, max depth refers to the longest path from the root node to a leaf node within each decision tree. This parameter controls how deep each tree can grow, allowing the model to capture varying levels of detail and complexity in the data. A larger max depth enables the trees to learn more intricate patterns, while a smaller max depth may yield simpler models that generalize better, potentially missing some nuances. Common max depth settings include 3, 5, or 7, and for this model, a depth of 4 strikes a balance between complexity and generalization.

The Random Forest model, an integrated ensemble learning approach, builds multiple decision trees and combines their outputs through a majority voting mechanism. This reduces variance and enhances robustness in early sepsis prediction. Each tree captures distinct aspects of the feature-sepsis relationship, and from these trees, a target decision tree path is identified. This path represents the sequence of feature splits that most accurately predicts the risk of sepsis based on the adjusted data. By utilizing this approach, the model provides a detailed understanding of the factors contributing to sepsis risk and offers valuable insights for effective intervention strategies.

## 2.5 Comparative Analysis of the COMPOSER and SXI++ LNM Models for Early Sepsis Detection:

The COMPOSER model (Conformal Multidimensional Prediction of Sepsis Risk) is a deep learning-based system designed for early sepsis detection. It incorporates a conformal prediction framework that handles uncertainty in predictions, effectively flagging cases where data is incomplete, erroneous, or deviates significantly from the training distribution. This mechanism minimizes false alarms and provides reliable early warnings.

**Use Cases**

1. **Early Sepsis Detection in ICUs and EDs**: COMPOSER provides early warnings of sepsis onset, enabling timely interventions. It identifies septic patients approximately **12.2 hours before ICU antibiotic administration** and **2.1 hours before ED interventions**.
2. **Clinical Workflow Integration:** It integrates seamlessly with clinical systems, offering actionable insights to clinicians through alerts and reducing unnecessary resource use.
3. **Improved Patient Outcomes:** A quasi-experimental study at UC San Diego demonstrated that COMPOSER led to a 1.9% reduction in in-hospital sepsis mortality and a 5.0% increase in sepsis bundle compliance.

**Limitations**
1. **Accuracy Limitations:** Although COMPOSER achieves high AUC values (ICU: 0.925–0.953; ED: 0.938–0.945), it falls short of the near-perfect performance achieved by SXI++ LNM.
2. **False Negative Risk:** While it minimizes false positives, the indeterminate class may include some septic patients, potentially delaying interventions.
3. **Adaptability:** COMPOSER's reliance on static conformal prediction may hinder its ability to handle high-dimensional or complex data as effectively as more dynamic systems like SXI++.

**Comparison with SXI++ LNM**
The **SXI++ Large Numerical Model (LNM)** employs a dynamic scoring system enhanced by a deep neural network, enabling iterative optimization of feature weights. In comparison:

- **Performance:** SXI++ consistently outperforms COMPOSER, achieving an AUC of 0.99–1.00 and precision/accuracy near 99.9%.
- **Generalization:** SXI++ handles unseen and diverse datasets effectively, maintaining robustness in real-world clinical settings.
- **Actionable Insights:** SXI++ provides detailed feature-level insights, enabling targeted interventions for sepsis management, which COMPOSER lacks.

**Specific Scenarios Where SXI++ LNM Outperforms COMPOSER**
1. **Imbalanced Datasets:** SXI++ addresses class imbalances using advanced weight adjustments, ensuring superior performance in datasets with low sepsis prevalence (e.g., 1.8%).
2. **Complex and High-Dimensional Data:** The SXI++ framework optimizes features dynamically, outperforming COMPOSER in settings with heterogeneous patient populations.
3. **Precision Critical Applications:** In scenarios demanding near-perfect accuracy to avoid misclassification, SXI++'s precision is unparalleled.
4. **Adaptability:** SXI++ excels in adapting to distributional shifts and unseen test data, while

COMPOSER relies more on its static conformal predictions.

## 3. Results:

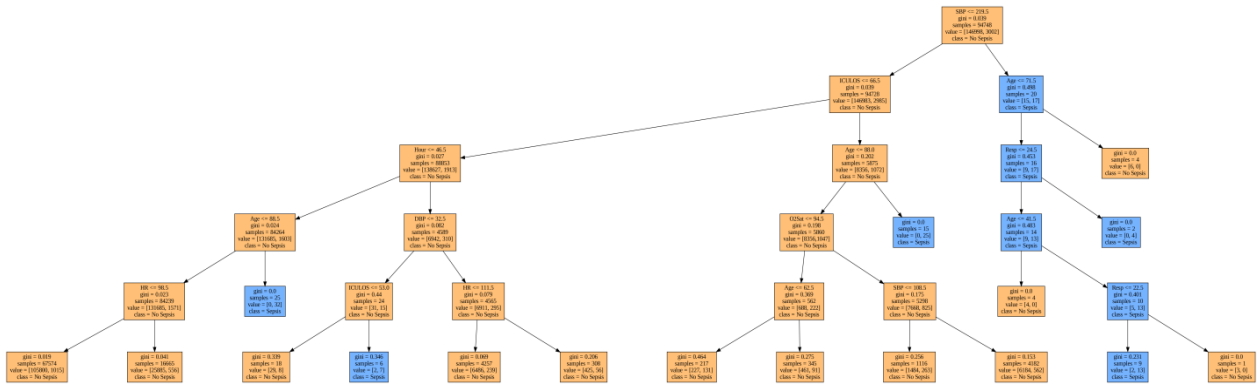

Fig: 3 Sepsis Detection Decision Tree Graph

From the tree in (Fig 3) we can conclude that (**pathways for early sepsis detection**): Patients likely to have sepsis should be detected within a 6-hour window if specific conditions are met. One critical indicator is a Systolic Blood Pressure (SBP) that exceeds 220, which may suggest a potential absence of sepsis. Additionally, if the patient falls within the age range of 42 to 72 years, this further increases the likelihood of sepsis. Another significant factor is the Respiration Rate; rates below 25 breaths per minute, or even lower than 23 breaths per minute, further indicate a higher probability of early sepsis.

Conversely, there are specific conditions that suggest a patient is unlikely to have sepsis. Other factors include a length of stay in the Intensive Care Unit (ICULOS) of less than 66.5 hours, an observation time of less than 46.5 hours, and a patient age below 89. Additionally, a Heart Rate (HR) of less than 98.5 beats per minute collectively supports the conclusion that the patient is unlikely to have sepsis.

Table 2: Model Performance Metrics for Case 1

| Metrics & Dataset Details | DATA | | LNM-1 (unseen data 1) (95% CI) | LNM-2 (unseen data 2) (95% CI) |
|---|---|---|---|---|
| | AUTO-ML | SXI++ | | |
| Accuracy | 77 | 99.99 (99.98 - 100.00) | 99.76 (99.72 - 99.80) | 99.99 (99.98–100.0) |
| Precision (PPV) | 78.45 | 99.79 (99.75 - 99.83) | 99.76 (99.72 - 99.80) | 99.9 (99.85–99.95) |
| Recall (NPV) | | 100 | 100 | 100 |
| AUC | 0.79 | 1 (0.99–1.00) | 0.99 (0.98 - 0.99) | 1 (0.99–1.00) |
| Number of rows: | 50404 (Sepsis: 15404, No Sepsis: 35000) | | 50000 (Sepsis: 1000, No Sepsis: 49000) | |
| Testing | 10000 | | | |
| Target variable distribution | No Sepsis: 69.4% Sepsis: 30.6% | | No Sepsis: 98% Sepsis: 2% | |

*Case 1 evaluates the SXI++ model on datasets with a higher proportion of sepsis cases, ensuring performance reliability.*

Table 3: Model Performance Metrics for Case 2

| | CASE - II | | | |
|---|---|---|---|---|
| Metrics | DATA | | LNM-1 (unseen | |

| Metrics & Dataset Details | AUTO-ML | SXI++ | LNM-1 (unseen data 1) (95% CI) | LNM-2 (unseen data 2) (95% CI) |
|---|---|---|---|---|
| Accuracy | 73.19 | 95.26 (95.07 - 95.45) | 90.12 (89.86 - 90.38) | 90.18 (89.90–90.46) |
| Precision (PPV) | 74.67 | 96.43 (96.26 - 96.60) | 92.81 (92.58 - 93.04) | 91.29 (91.00–91.58) |
| Recall (NPV) | | 98.14 (97.87-98.41) | 96.64 (96.48-96.81) | 94.91 (94.71- 95.11) |
| AUC | 0.75 | 0.92 (0.91 - 0.93) | 0.90 (0.897 - 0.903) | 0.90 (0.91–0.93) |
| Number of rows: | 50,000 (Sepsis: 1000, No Sepsis: 49000) | | 50,000(Sepsis: 1000, No Sepsis: 49000) | |
| Testing | 10,000 | | | |
| Target variable distribution | No Sepsis: 98% Sepsis: 2% | | | |

*Case 2 examines model performance on highly imbalanced datasets, where detecting rare sepsis cases is critical.*

Table 4: Model Performance Metrics for Case 3

| CASE - III | | | | |
|---|---|---|---|---|
| Metrics & Dataset Details | DATA | | LNM-1 (unseen data 1) (95% CI) | LNM-2 (unseen data 2) (95% CI) |
| | AUTO-ML | SXI++ | | |
| Accuracy | 84.09 | 99.9 (99.84 - 99.96) | 97.78 (97.49 - 98.07) | 97.89 (97.70–98.08) |
| Precision (PPV) | 86.19 | 99.9 (99.84 - 99.96) | 99.04 (98.85 - 99.23) | 99.21 (99.05–99.37) |
| Recall (NPV) | | 99.93 (99.926-99.934) | 98.67 (98.66 - 98.68) | 98.56 (98.55 -98.57) |
| AUC | 0.87 | 1 (1.00 -1.00) | 0.98 (0.97 -0.981) | 0.99 (0.98–0.99) |
| Number of rows: | 10000 (Sepsis: 3000, No Sepsis: 7000) | | 50000(Sepsis: 900, No Sepsis: 49100) | |
| Testing | 2,000 | | | |
| Target variable distribution | No Sepsis: 70% Sepsis: 30% | | No Sepsis: 98.2% Sepsis: 1.8% | |

*Case 3 evaluates model robustness with a balanced training dataset and an imbalanced test dataset to simulate real-world clinical scenarios.*

Above tables presents the evaluation of the SXI++ LNM model across three distinct cases, each designed to test its robustness under varying data distributions. CASE-1 (Table 2) utilizes a larger training dataset with a higher proportion of sepsis cases (30.6%) and evaluates performance on an imbalanced test set (98% non-sepsis, 2% sepsis). CASE-2 (Table 3) examines model performance when both training and test datasets are highly imbalanced (98% non-sepsis, 2% sepsis), assessing its ability to detect rare sepsis cases. CASE-3 (Table 4) investigates generalization by training on a relatively balanced dataset (30% sepsis cases) and testing on an imbalanced set (1.8% sepsis cases), ensuring the model can adapt to real-world clinical distributions. These cases collectively validate the model's ability to handle class imbalances and maintain high predictive accuracy across diverse scenarios.

In above tables, Auto-ML or automated machine learning refers to an automatically constructed and executed pipeline for machine tasks (23). COMPOSER refers to the results presented in the COMPOSER papers (24, 25) on the same dataset we use, allowing for direct comparison in the table. The AUC of 0.99 shown in (Table 5) indicates that SXI++ consistently maintains high performance across different scenarios. Overall, these results establish SXI++ as a highly reliable tool for early sepsis detection, providing clinicians with confidence in its predictive abilities and facilitating timely interventions that could improve patient outcomes.

Table: 5 AUC Comparison

| Metric | COMPOSER (95% CI) | SXI++ LNM (95% CI) |
|---|---|---|
| AUC | 0.91 - 0.95 (0.89–0.96) | 0.99 (0.98–1.00) |

Composer & SXI++LNM with 95% Confidence Interval

SXI++ LNM employs an advanced deep neural network framework that incorporates dynamic weight adjustment and hyperparameter tuning, significantly enhancing its predictive power for early sepsis detection. This sophisticated architecture enables SXI++ LNM to efficiently process large datasets while iteratively calibrating weights, which contributes to its superior performance metrics. The strategic improvement of SXI++ scores is crucial, as higher scores are associated with lower sepsis rates. This emphasizes the importance of focusing on enhancing SXI++ scores to achieve better health outcomes in sepsis management.

In the comparison between SXI++ LNM and the COMPOSER algorithm for early sepsis detection within a critical six-hour window, both models exhibit high effectiveness but differ in their strengths and applications. SXI++ LNM stands out with exceptional performance metrics, often approaching 99% accuracy and high AUC values, indicating its robust capability to accurately distinguish between septic and non-septic patients with minimal errors. Its high precision further underscores its reliability in predicting early sepsis onset accurately. Conversely, while COMPOSER integrates a conformal prediction mechanism aimed at reducing false alarms, its AUC values—though impressive, ranging from 0.91 to 0.94—do not reach the near-perfect accuracy of SXI++. Furthermore, SXI++ LNM consistently maintains high accuracy and precision, even with unseen data, making it a powerful tool for early sepsis detection. On the other hand, COMPOSER's strength lies in its ability to generalize across diverse patient populations and clinical environments, making it suitable for real-world applications where variations in patient demographics and clinical practices occur. Overall, both models play essential roles in early sepsis detection, but SXI++ LNM demonstrates superior capabilities in precision and accuracy the strategic improvement of SXI++ scores is crucial.

**Discussions**

**Key Findings**

This study revealed that the SXI++ Large Numerical Model (LNM) significantly outperformed the COMPOSER algorithm in early sepsis detection, achieving near-perfect accuracy (up to 99.99%), precision (99.9%), and AUC (0.99–1.00). These findings align with prior studies emphasizing the importance of integrating advanced machine learning algorithms in sepsis prediction. Unlike static models, SXI++ demonstrated adaptability across diverse datasets by employing a dynamic scoring system and deep learning optimization, offering actionable insights for clinicians.

**Strengths and Limitations**

The use of a large, representative dataset from three distinct U.S. hospital systems is a key strength of this study, ensuring generalizability and real-world applicability. However, limitations exist. COMPOSER's reliance on conformal predictions can reduce false alarms but may underperform in handling complex data relationships. Moreover, the SXI++ model has yet to be validated on external, non-U.S. datasets or in non-ICU settings, which limits its applicability. Additionally, the lack of real-time implementation and external evaluations constrains its immediate clinical utility.

The current use of SIRS to diagnosis sepsis is widely associated with poor specificity. Patient monitoring software systems such as the Epic EHR Sepsis Predictive Analytic Tool have very low PPV (Positive predictive value or Precision) of 33% resulting in many erroneous alarms. These alarms result in additional work for staff and contribute to alarm fatigue. Our proposed work with SXI++ provides predictive analytics with high accuracy that exceeds the current state-of-the-art, including COMPOSER.

**Comparison with Similar Research**

The SXI++ model builds on prior machine learning advancements in sepsis detection, improving upon models such as TREWScore and PhysioNet Challenge solutions. These models, while effective in specific scenarios, lacked the dynamic scoring and iterative optimization capabilities of SXI++. For example, TREWScore identified sepsis within 6 hours but had limitations in precision when applied to heterogeneous populations. COMPOSER's AUC values (0.91–0.95) are commendable but fall short of SXI++'s superior metrics, especially in handling imbalanced datasets where positive cases are scarce.

**Explanations of Findings**

The superior performance of SXI++ is rooted in its use of a composite scoring system that integrates multiple machine learning algorithms, optimizing feature weights iteratively through a proprietary deep neural network. This adaptability enables it to generalize well across diverse datasets, a challenge for static models like COMPOSER. COMPOSER's reliance on conformal prediction limits its capability to handle shifts in data distribution and high-dimensional inputs, which are better addressed by SXI++'s robust architecture.

**Implications, Recommendations, and User Interaction**

Implications: The SXI++ LNM has the potential to transform critical care practices by enabling earlier and more precise sepsis detection. Its adaptability across datasets suggests it could be effectively implemented in diverse clinical environments, improving outcomes and resource allocation.

Recommendations: To enhance its clinical adoption, future research should:

- Validate SXI++ on external datasets, including non-U.S. and outpatient populations.
- Integrate SXI++ into electronic health record systems for real-time predictions.
- Conduct prospective trials to evaluate its impact on patient outcomes.

User Interaction and Expertise: The SXI++ LNM requires minimal user interaction during data handling, as it automates preprocessing and feature selection. Users need to provide structured input data (e.g., vital signs, lab results) and define the target outcome. While no specialized expertise in machine learning is required, basic knowledge of clinical data and predictive analytics will enhance its effective use. An intuitive interface could further simplify adoption by clinicians and administrative staff.

**Future Investigation:**

Future investigations include studies on antibiotic choice, design of integrating this technology into software platforms conforming to most used HL7 FHIR interoperability standards and HIPAA-secure data standards, and further validation. These points are further developed in this section.

Future research should prioritize validating the SXI++ LNM model on external, non-U.S. datasets and non-ICU settings to confirm its generalizability across diverse healthcare environments. Prospective clinical trials are essential to assess its real-world impact on patient outcomes, including mortality reduction and adherence to sepsis management protocols. Integration into electronic health record systems, aligned with HL7 FHIR interoperability and HIPAA-secure standards, would facilitate real-time clinical implementation. Additionally, exploring its utility in guiding antibiotic stewardship and optimizing resource allocation could further enhance its clinical relevance. Addressing these steps will bridge current gaps in adaptability, external validation, and practical deployment, ensuring SXI++ LNM's potential to transform sepsis care globally.

## 4. Conclusions

In this study, we thoroughly explored the efficacy of the SXI++ LNM framework for early sepsis detection, a critical area of concern in healthcare where timely intervention can significantly impact patient outcomes. By integrating advanced deep neural network methodologies with dynamic weight adjustment and hyperparameter tuning, SXI++ LNM demonstrated superior predictive capabilities compared to traditional models. The framework effectively handles large datasets and employs iterative calibration processes, allowing for robust performance metrics that approach 99% accuracy and exceptional precision in identifying sepsis cases. Our findings highlighted the correlation between enhanced SXI++ scores and reduced sepsis rates, underscoring the importance of improving these scores to achieve better health outcomes. The capability of the SXI++ LNM to perform effectively across various patient populations and settings was also validated. This adaptability is crucial for real-world applications, where patient demographics and clinical practices can vary significantly, ensuring that the model remains relevant and reliable in diverse healthcare environments.

The performance of the SXI++ LNM was further assessed against the COMPOSER algorithm, showcasing the distinct strengths of each model. While COMPOSER excels in reducing false alarms through a comprehensive conformal prediction mechanism, its AUC values, though impressive, do not rival the near-perfect accuracy of the SXI++ LNM. This distinction highlights SXI++ LNM's robust capability in accurately predicting early sepsis onset, making it an invaluable tool for clinicians in urgent settings. In addition to performance metrics, we conducted comparative analyses using various unseen datasets, further demonstrating the consistency of the SXI++ LNM's high accuracy and precision. The model's ability to maintain superior performance across different data scenarios underscores its reliability in critical applications, emphasizing the need for healthcare providers to consider adopting such advanced predictive technologies.

**Acknowledgments**

None

**Footnote:**

*Reporting Checklist:* The authors have completed the TRIPOD reporting checklist. Available at https://jmai.amegroups.com/article/view/10.21037/jmai-24-393/rc

*Peer Review File:* Available at https://jmai.amegroups.com/article/view/10.21037/jmai-24-393/prf

*Funding:* None.

*Conflicts of Interest:* All authors have completed the ICMJE uniform disclosure form (available at https://jmai.amegroups.com/article/view/10.21037/jmai-24-393/rc). All authors are employed by Sriya.AI LLC. Sriya.AI LLC has filed US provisional patents on the underlying core technology and its applications in the healthcare industry. 4 patents related to the technology is already filed which is being cited in references (19, 20, 21, 22). The authors have no other conflicts of interest to declare.

*Ethical Statement:* The authors are accountable for all aspects of the work in ensuring that questions related to the accuracy or integrity of any part of the work are appropriately investigated and resolved. Since the research involved secondary analysis of anonymized data and did not include direct interaction with or intervention in human subjects, an ethics board review was not required. The study did not involve sensitive or personal data that would necessitate informed consent.